\documentclass[conference]{IEEEtran}
\IEEEoverridecommandlockouts
\usepackage{cite}
\usepackage{amsmath,amssymb,amsfonts}
\usepackage{algorithm}
\usepackage{algpseudocode}
\usepackage{graphicx}
\usepackage{caption}
\usepackage[normalem]{ulem}
\usepackage{ragged2e}
\usepackage{siunitx}
\usepackage{multirow}
\usepackage{subfigure}
\usepackage{subcaption}
\usepackage{textcomp}
\usepackage{url,comment}
\newcommand{\authtwo}[1]{\textcolor{black}{#1}}  

\usepackage{xcolor}
\def\BibTeX{{\rm B\kern-.05em{\sc i\kern-.025em b}\kern-.08em
    T\kern-.1667em\lower.7ex\hbox{E}\kern-.125emX}}
    
\begin{document}

\title{Securing Traffic Sign Recognition Systems in Autonomous Vehicles}
\author{
\IEEEauthorblockN{Thushari Hapuarachchi, Long Dang, Kaiqi Xiong}
\IEEEauthorblockA{ICNS Lab and Cyber Florida, \\ University of South Florida,
Tampa, FL, 33620 USA\\
Email: saumya2@usf.edu, longdang@usf.edu, xiongk@usf.edu}
}

\maketitle

\begin{abstract}
Deep Neural Networks (DNNs) are widely used for traffic sign recognition because they can automatically extract high-level features from images. These DNNs are trained on large-scale datasets obtained from unknown sources. Therefore, it is important to ensure that the models remain secure and are not compromised or poisoned during training. In this paper, we investigate the robustness of DNNs trained for traffic sign recognition. First, we perform the error-minimizing attacks on DNNs used for traffic sign recognition by adding imperceptible perturbations on training data. Then, we propose a data augmentation-based training method to mitigate the error-minimizing attacks. The proposed training method utilizes nonlinear transformations to disrupt the perturbations and improve the model robustness. We experiment with two well-known traffic sign datasets to demonstrate the severity of the attack and the effectiveness of our mitigation scheme. The error-minimizing attacks reduce the prediction accuracy of the DNNs from 99.90\% to 10.6\%. However, our mitigation scheme successfully restores the prediction accuracy to 96.05\%. Moreover, our approach outperforms adversarial training in mitigating the error-minimizing attacks. Furthermore, we propose a detection model capable of identifying poisoned data even when the perturbations are imperceptible to human inspection.  Our detection model achieves a success rate of over 99\% in identifying the attack. This research highlights the need to employ advanced training methods for DNNs in traffic sign recognition systems to mitigate the effects of data poisoning attacks.

\end{abstract}


\section{Introduction}
Traffic sign recognition systems are essential for detecting road signs and aiding drivers or control modules in making informed driving decisions. Modern autonomous vehicle models, such as Tesla’s Model 3~\cite{tesla}, integrate traffic sign recognition systems as essential components of their driving assistance technology\cite{DBLP:journals/corr/abs-2409-04133}.
According to~\cite{DBLP:journals/sensors/LimLLOAA23}, the classification algorithms employed in traffic sign recognition systems can be categorized into two groups: machine learning-based and deep learning-based approaches. Deep learning algorithms include deep neural networks (DNNs) which are widely used nowadays because they can automatically extract high-level features from images. However, researchers have expressed concerns about the security of traffic sign recognition systems, as their dependence on DNNs makes them susceptible to various attacks such as data poisoning~\cite{DBLP:journals/tvt/JiangLLLL20, LinMLAttacks}.

\begin{figure}
    \centering
    \includegraphics[scale =0.55]{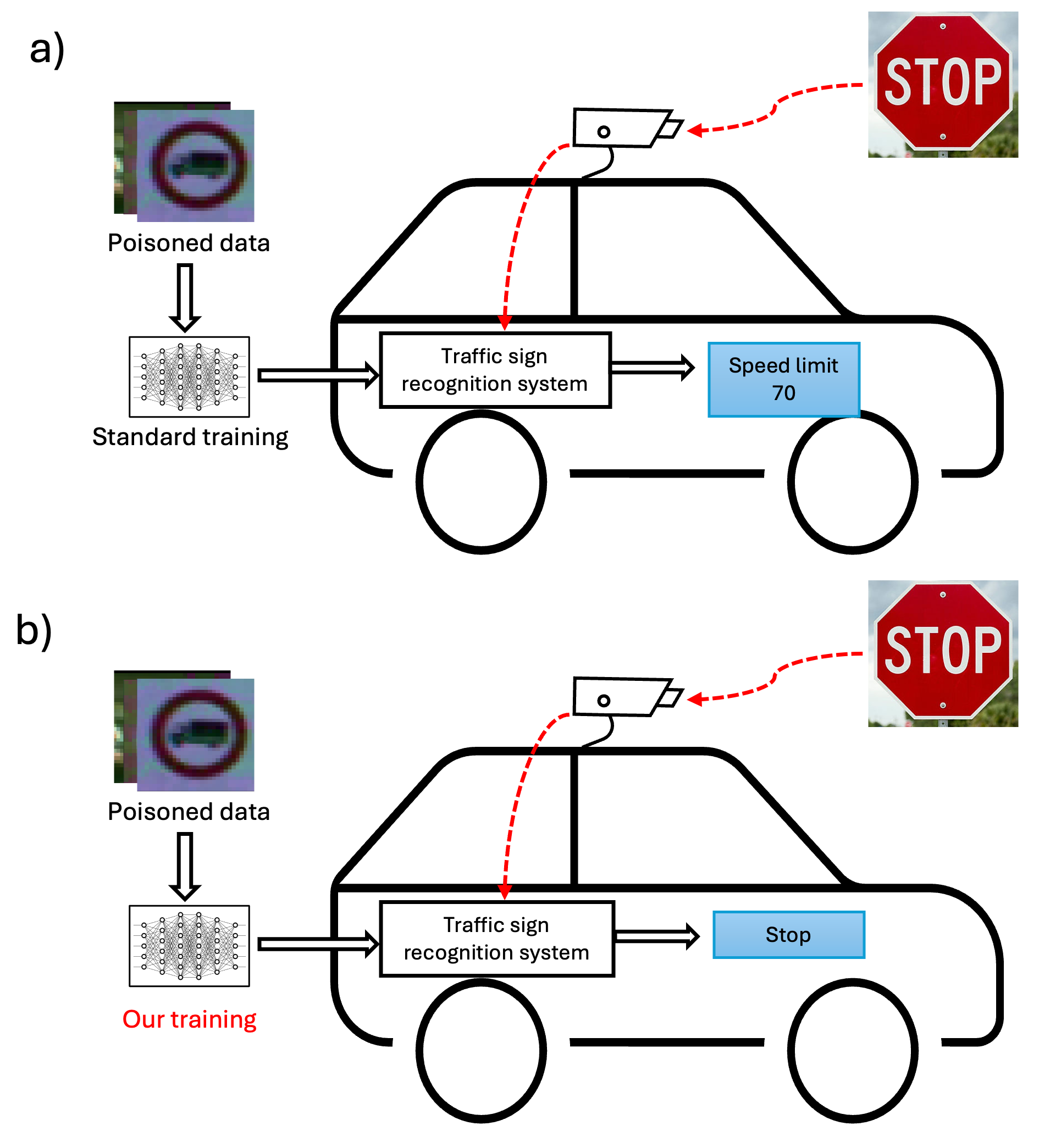}
    \caption{Overview of this research. a) We train the traffic sign recognition system using data poisoned by error-minimizing attacks with \textit{standard training}. When the trained system is used for predicting traffic signs on the road, the signs are misclassified. b) We train the traffic sign recognition system with \textit{our mitigation scheme}, where we propose a data-augmentation-based training method to mitigate the effects of  error-minimizing attacks. Then, the trained traffic sign recognition system is able to provide the correct predictions. }
    \vspace{-0.4cm}
    \label{fig:overview}
    
\end{figure}

DNNs for traffic sign recognition systems need a lot of data for training to make sure that they perform without any errors that might cause fatal accidents. As a result, data may be collected from various sources, both trusted and untrusted, such as the Internet, for training purposes. However, these data may be poisoned. One such method of data poisoning is the error-minimizing attacks~\cite{DBLP:conf/iclr/HuangME0021}, which involve adding imperceptible perturbations to the data. These data are difficult to be detected as poisoned due to the low intensity of the perturbations. Once the system is trained on these data, the training accuracy is very high. However, when the system predicts images captured by the vehicle on the road, the prediction accuracy~\cite{DBLP:journals/tvt/JiangLLLL20}  drops, leading to incorrect predictions. Fig.~\ref{fig:overview}-a) illustrates this scenario.

In this paper, we first implement \authtwo{the} error-minimizing attacks~\cite{DBLP:conf/iclr/HuangME0021} on DNNs trained for traffic sign recognition. We manipulate the strength of the attack by varying the perturbation intensity. The strength of the attack is measured using the prediction accuracy. Our experimental results show that the attack is stronger when the perturbation intensity is higher. However, the perturbations become visible at higher intensities and can be easily detected by data collectors. To detect this attack, we propose a detection model capable of identifying poisoned data even when the perturbations are imperceptible to human inspection.  The detection model is built using a simple \authtwo{convolutional neural network} (CNN) model and gave \authtwo{a success rate of more than 99\%}. Moreover, we propose a data augmentation-based training method (a mitigation scheme) to mitigate error-minimizing attacks, as illustrated in Fig.~\ref{fig:overview}-b). The proposed training method utilizes nonlinear transformations such as grayscale to disrupt the perturbations and improve the model's resilience.  This mitigation scheme is able to achieve almost the same performance as a system trained on clean data. Furthermore, we show that the error-minimizing attack has a limitation: when DNNs are exposed to clean (non-poisoned) data during training, the attack becomes ineffective.

\authtwo{The} error-minimizing attacks, as introduced by~\cite{DBLP:conf/iclr/HuangME0021},  \authtwo{have} not been implemented on traffic sign recognition datasets. When applying \authtwo{the} error-minimizing attacks on traffic sign recognition datasets, we face several challenges compared to \authtwo{the baseline datasets} \authtwo{including CIFAR-10, and ImageNet}:
1) handling complex datasets that \authtwo{have} larger number of classes, 2) limited variety of images within the same class, 3) high variations in image colors in the same class, and 4) algorithms require longer convergence times. We explain how we addressed these challenges in Sec.~\ref{sec.method_attack}. Additionally, we compare the proposed mitigation scheme with adversarial training, a widely used approach for mitigating evasion attacks~\cite{madry2017AT_PGD}. The experimental results demonstrate that our mitigation scheme outperforms adversarial training. 
In summary, we make the following \authtwo{key} contributions.

\begin{itemize}
\vspace{-0.04cm}
    \item We exploit \authtwo{the} error-minimizing attacks to poison DNNs in traffic sign recognition systems. The attacks significantly drop the prediction accuracy of the traffic sign recognition systems.
    \item We propose a detection model to identify the data poisoned by \authtwo{the} error-minimizing attacks. The detection model is  \authtwo{built} using a simple CNN model and resulted \authtwo{in a success rate of more than 99\%}.
    \item We propose a data augmentation-based training method (a mitigation scheme) to mitigate \authtwo{the} effect of error-minimizing attacks. The proposed training method utilizes nonlinear transformations to disrupt the perturbations added by \authtwo{the} error-minimizing attacks. Our experimental results show that the proposed training method outperforms adversarial training in mitigating the error-minimizing
attacks.
\end{itemize}

\section{Related Work}
DNNs are susceptible to both evasion and data poisoning attacks, depending on the stage at which the attack is executed~\cite{10863718,DBLP:journals/tvt/JiangLLLL20, LinMLAttacks}. Data poisoning attacks involve injecting malicious or misleading samples into the training dataset, ultimately degrading the model's generalization and reducing its prediction accuracy. These attacks can be particularly stealthy, as poisoned samples may appear benign but cause systematic misclassifications. On the other hand, evasion attacks occur at prediction time by adding carefully crafted adversarial perturbations to input images, tricking trained DNNs into misclassifying them~\cite{DBLP:conf/ccs/PapernotMGJCS17,DBLP:conf/ccs/01900LLHZ21}. Such attacks pose significant risks in real-world applications, especially for autonomous driving systems that rely heavily on accurate traffic sign recognition~\cite{10863718,DBLP:journals/vcomm/LimbasiyaTCZ22}.

In this paper, we explore the error-minimizing attacks~\cite{DBLP:conf/iclr/HuangME0021} to attack DNNs trained for traffic sign recognition. This attack strategically perturbs training samples to maximize the model training accuracy while minimizing the model's prediction ability on legitimate (clean) data. Several studies~\cite{Thushari24DA,liu2021going,DBLP:conf/icml/Liu0L23} have proposed remedies for \authtwo{the} error-minimizing attacks including data augmentation techniques and adversarial training~\cite{DBLP:conf/iclr/FuHLST22}. However, these studies mostly focused on \authtwo{the} baseline datasets. In contrast, our study highlights the effectiveness of this attack in traffic sign recognition systems and proposes a mitigation scheme incorporating nonlinear transformations to reduce the impact of \authtwo{the} error-minimizing attacks.


\section{Methodology}
\label{sec.method}

\subsection{Error-minimizing Attacks} \label{sec.method_attack}
\authtwo{The} error-minimizing attacks~\cite{DBLP:conf/iclr/HuangME0021} make poisoned data by adding a type of imperceptible perturbation to the original data. This perturbation is generated in a way that minimizes the loss of the model trained on it while preventing the DNN from learning the actual features of the images. 
In this study, we adopted Huang et al.~\cite{DBLP:conf/iclr/HuangME0021}'s approach for performing \authtwo{the} error-minimizing attacks. They proposed solving the following bi-level optimization problem to generate the perturbations.
\begin{equation}
\label{eq:emin}
    \arg\min_\theta\left[ \min_{\|\delta\|_p \leq \epsilon} \mathcal{L}(f(X^D+\delta;\theta),Y^D) \right],
\end{equation}
where $X^D$ and $Y^D$ denote a set of training images and a set of target labels, respectively. The goal of the inner optimization problem is to find the set of perturbations denoted by $\delta$.  Huang et al.~\cite{DBLP:conf/iclr/HuangME0021} used the Projected Gradient Descent (PGD) method to solve this inner optimization problem. First, they initialized  $\delta$.  In each iteration, a fixed model with parameters $\theta$ is evaluated on $X^D + \delta$ and the loss is calculated. $\delta$ is updated for $T$ iterations while minimizing the loss. $\epsilon$ controls the intensity of the added perturbation. The $L_p$ norm of $\delta$ is bounded by $\epsilon$ to ensure that the perturbations remain imperceptible. 
The outer optimization problem finds the model parameters that minimize the same loss. In this step, $\delta$ is fixed and the model parameters are optimized to achieve the minimum loss when trained on $X^D + \delta$.  The optimization process is terminated when the model's training accuracy is higher than a given value $\lambda$.  

We implemented this attack on the traffic sign recognition system, an aspect \authtwo{that was} not explored by Huang et al.~\cite{DBLP:conf/iclr/HuangME0021}. Applying this attack on traffic sign data presents greater challenges compared to \authtwo{the} baseline datasets due to several factors. 1) The higher number of classes in the traffic sign datasets compared to the baseline datasets increases the complexity of the perturbation generation process. The datasets we considered, German Traffic Sign Recognition Benchmark (GTSRB)~\cite{Houben-IJCNN-2013} and the Chinese Traffic Sign Recognition Database (CTSRD)~\cite{tsrd}, \authtwo{have} 43 and 58 classes, respectively. 2) The limited variety of images within the same class makes it easier for models to learn them effectively, which makes the attacks more difficult. To address this, we generate class-wise perturbations instead of sample-wise perturbations. 3) Variations in image appearance, such as differences between night and day, add another layer of difficulty. Some images are particularly dark, making it more difficult to introduce imperceptible perturbations. 4) Due to the higher complexity of the data, the algorithms require longer convergence times. Hence, we set the number of PGD iterations to solve the inner minimization problem in Eq.~\ref{eq:emin} to 1 instead of 10 which is suggested by \cite{DBLP:conf/iclr/HuangME0021}.


In this study, we used ResNet18 as the model architecture when generating the attack. We chose ResNet18  \authtwo{because} is it a widely used high performance DNNs model for image classification~\cite{DBLP:conf/cvpr/HeZRS16}. We consider\authtwo{ed} a white-box attack setting and attack\authtwo{ed} the same model. The stopping criterion for generating perturbations  \authtwo{was} set to 99\% ($\lambda$) training accuracy. 
We consider\authtwo{ed} three values for $\epsilon$. We define\authtwo{d} the strength of the attack based on $\epsilon$, as the attack's performance highly depends on the intensity of the added perturbations.

\vspace*{-0.1cm}
\subsection{The Detection Model}
\begin{figure}[ht!]
    \centering
    \includegraphics[scale =0.35]{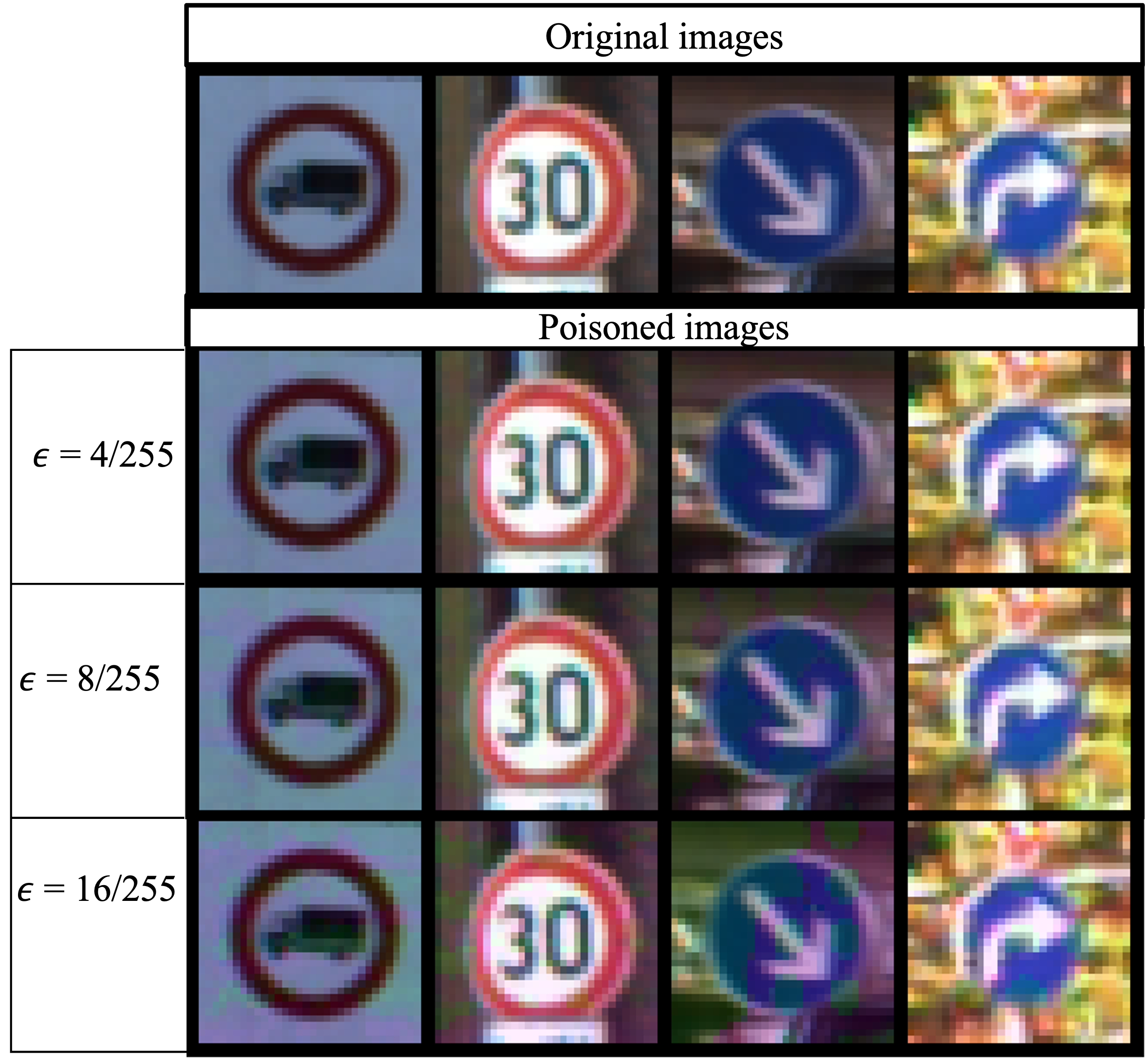}
    \caption{Original images and poisoned images with different poison rates.}
    \label{fig:examples}
    \vspace*{-0.4cm}
\end{figure}

As shown in Fig~\ref{fig:examples}, \authtwo{the} error-minimizing attacks perturbations are barely visible to the human eyes, especially when $\epsilon =4$. Hence, model trainers are unable to determine whether the data is poisoned by merely observing the images. Additionally, it is not feasible to inspect all images manually. Therefore, we proposed a detection model to identify whether each image is poisoned or clean. The output of the detection model is either `poisoned' or `clean'. 

\begin{figure}[ht!]
    \centering
    \includegraphics[width=\linewidth]{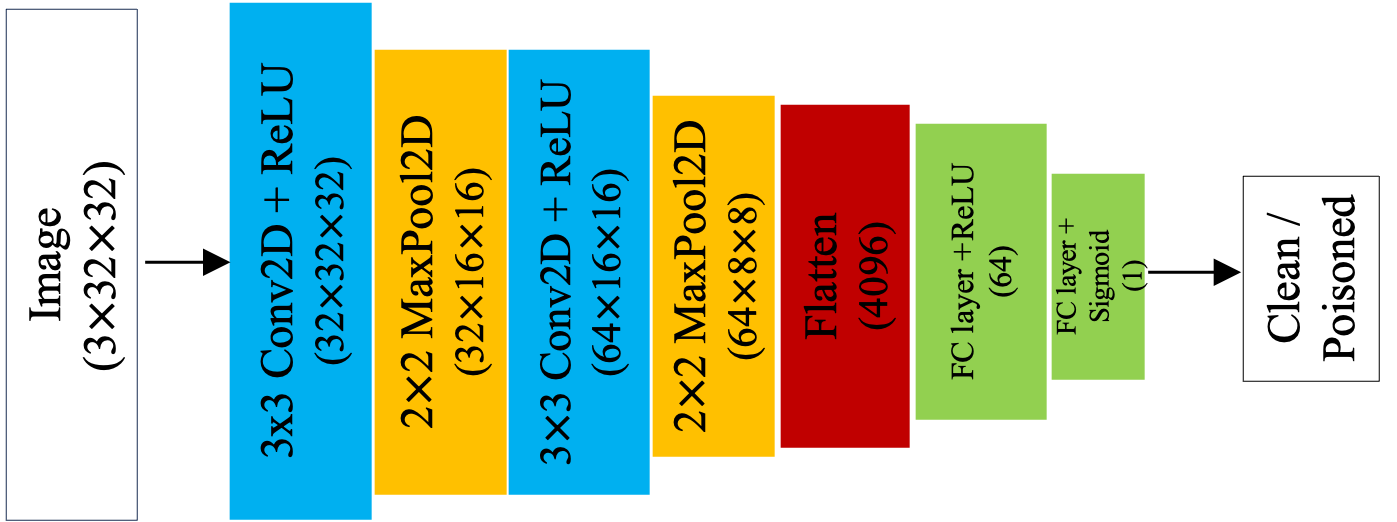}
    \caption{The detection model.}
    \label{fig:detectmod}
    \vspace*{-0.4cm}
\end{figure}

Fig~\ref{fig:detectmod} shows the model architecture of the detection model. The model consisted of two convolutional layers with a kernel size of 3$\times$3 followed by a max pooling layer with a kernel size of 2$\times$2, flattening, and two fully connected layers. The output size of each layer is shown in Fig~\ref{fig:detectmod}.  The final layer used a sigmoid activation function to classify images into two categories i.e, poisoned or clean image. All other layers \authtwo{have} ReLU activation function. The detection model utilized binary cross-entropy loss as the loss function and Adam as the optimizer.

\subsection{The Mitigation Scheme}
The goal of the mitigation scheme is to recover the traffic sign recognition system that has been attacked by \authtwo{the} error-minimizing attacks. To achieve this, we proposed a data augmentation-based training approach instead of the standard training method. Algorithm~\ref{alg:Method} shows the proposed mitigation scheme. First, we select\authtwo{ed} a set of nonlinear transformation techniques ($T$). When selecting $T$, we prioritize\authtwo{d} the nonlinear transformations explored in previous studies for mitigating \authtwo{the} error-minimizing attacks~\cite{Thushari24DA,liu2021going,DBLP:conf/icml/Liu0L23}. Then, we randomly chose one transformation ($t_i$) from the set and applied it on the training dataset. Next, we trained the traffic sign recognition system using the transformed data. We chose ResNet18 as the model architecture and trained it for 20 epochs with the Stochastic Gradient Descent (SGD) optimizer and cross-entropy loss as the loss function. After training, we validated the model on a clean dataset, which reflected how the system performs on traffic sign images captured by the vehicle while driving. We defined the accuracy that is obtained on the clean validation dataset as \textit{prediction accuracy}. If the prediction accuracy reached the desired value ($\alpha$), we stopped the training process. Otherwise, we chose another transformation, applied it to the training dataset, and combined the newly transformed dataset with the dataset from the previous iteration. In this way, we expanded the training dataset size in each iteration until we achieve the desired prediction accuracy.

\begin{algorithm}[htb!]
  \caption{Mitigating error-minimizing attacks}
  \label{alg:Test}
  \begin{algorithmic}
    \State \textbf{Input:} Poisoned Traffic sign data ($D$), Traffic Sign recognition system ($F$), Clean prediction dataset ($C$), A set of nonlinear transformation techniques ($T$), Target prediction accuracy ($\alpha$)
    \State \textbf{Return:} Robust Traffic Sign recognition system ($F_i$)
    \While{$v < \alpha$}
        \State 1. Randomly choose $t_i$ from  $T$
        \State 2. Transform $D$ using $t_i$  $\rightarrow$ $t_i(P)$
        \State 3. $t_i(P)$ + $D_{i-1}$ $\rightarrow$ $D_i$ \Comment{$D_0 = \emptyset$}
        \State 4. Train $F$ on $D_i$ $\rightarrow$ $F_i$
        \State 5. Predict  $F_i$ on $C$ $\rightarrow$ prediction accuracy, $v$
    \EndWhile
  \end{algorithmic}
  \label{alg:Method}
\end{algorithm}

We evaluated our mitigation scheme on two traffic sign image datasets: GTSRB~\cite{Houben-IJCNN-2013}, and CTSRD~\cite{tsrd}. In our experimental evaluation, we chose three nonlinear transformations for $T$: grayscale, Color Jitter, and Random Invert. To implement these transformations, we used PyTorch's built-in options~\cite{transformspytorch}. For instance, we used the \textit{transforms.Grayscale(3)} command to apply the grayscale transformation, where 3 represents the number of channels we want the transformed image to have. We chose grayscale because, several studies~\cite{Thushari24DA,liu2021going,DBLP:conf/icml/Liu0L23}  have already shown that grayscale transformation is effective against \authtwo{the} error-minimizing attacks. Moreover, grayscale transformation controls the channel-wise perturbations by replacing all three channel values in a pixel with a single value. For the Color Jitter transformation, we used the command \textit{transforms.ColorJitter(brightness=.5, hue=.3)}, which reduced the brightness and hue of the image. By modifying the brightness and hue, the color jitter transformation expands the feature space of the model and reduces the risk of overfitting to adversarial examples~\cite{xiao2024exploring}. To apply the Random Invert transformation, we used the command \textit{transforms.RandomInvert(p=1.0)}, where $p=1.0$ indicates that the transformation is applied to all the images in the dataset. This transformation modifies the pixel values of an image by inverting them.

\section{Experimental Evaluation}

For the experimental evaluation, we used two datasets: the German Traffic Sign Recognition Benchmark (GTSRB)~\cite{Houben-IJCNN-2013}, and the Chinese Traffic Sign Recognition Database (CTSRD)~\cite{tsrd}. Table~\ref{tab:datasets} includes the specifications of these datasets. GTSRB dataset contains 31,367 images for training and 7,842 images for prediction (aka validation), spanning 43 classes. The CTSRD dataset includes 3,336 training images and 834 prediction images, covering 58 classes. All the experiments are conducted in PyTorch 1.13.1 framework.

\begin{table}[ht]
    \centering
    \vspace*{-0.2cm}
    \caption{Dataset details.}
    \begin{tabular}{|c|c|c|c|}
    \hline
    Dataset & Training set size & Prediction set size & \# of classes \\
    \hline
       GTSRB~\cite{Houben-IJCNN-2013}  & 31367 & 7842 & 43 \\
       \hline
       CTSRD~\cite{tsrd} & 3336 & 834 & 58 \\
    \hline
    \end{tabular}
    \label{tab:datasets}
    \vspace*{-0.4cm}
\end{table}

\subsection{Error-minimizing Attacks on Traffic Sign Recognition Systems}

To perform the error-minimizing attacks, we adopted the code in Huang et al.~\cite{DBLP:conf/iclr/HuangME0021}. ResNet18  \authtwo{was} used as the base model for generating the error-minimizing perturbations.  In our experiments, we fixed the seed to 42 to ensure the reproducibility. We generated class-wise perturbations, as they are more effective than sample-wise perturbations~\cite{DBLP:conf/iclr/HuangME0021}. Perturbations were generated with three maximum perturbation limits ($\epsilon$) of 4/255, 8/255, and 16/255. When $\epsilon = 4/255$, the perturbations are barely visible, making it difficult for human eyes to distinguish poisoned images (see Fig.~\ref{fig:examples}).  When $\epsilon = 16/255$, the perturbations became slightly visible. Hence, we defined the strength of the attack based on the $\epsilon$: a higher $\epsilon$ indicates a stronger attack due to the increased intensity of the perturbations. 
Following the white-box attack settings, we attacked the same model (ResNet18) using the poisoned data.

The second column in Table~\ref{tab:results} shows the prediction accuracy of the model that is trained on clean data (no attack). The prediction dataset is also clean because it reflects the images captured by the vehicle while driving on the road. The model achieved prediction accuracies of 99.90\% and 98.56\% for the GTSRB and CTSRD datasets, respectively. The fourth column of Table~\ref{tab:results} shows the accuracy of the prediction after the error-minimizing attacks. The attack is able to reduce the prediction accuracy  from 99.90\% to 10.6\% for the GTSRB dataset and  from 98.56\% to 25.18\% for the CTSRD. These results indicate that the poisoned models failed to accurately predict the signs in the traffic sign images. However, the strength of the attack can be changed by modifying $\epsilon$. When $\epsilon$ is reduced to $4/255$, the prediction accuracy increased to 52.10\% for GTSRB dataset and 70.26\% for CTSRD dataset. These results demonstrate that the attack becomes weaker as $\epsilon$ decreases.

\begin{figure}[ht!]
    \centering
    \includegraphics[width=\linewidth]{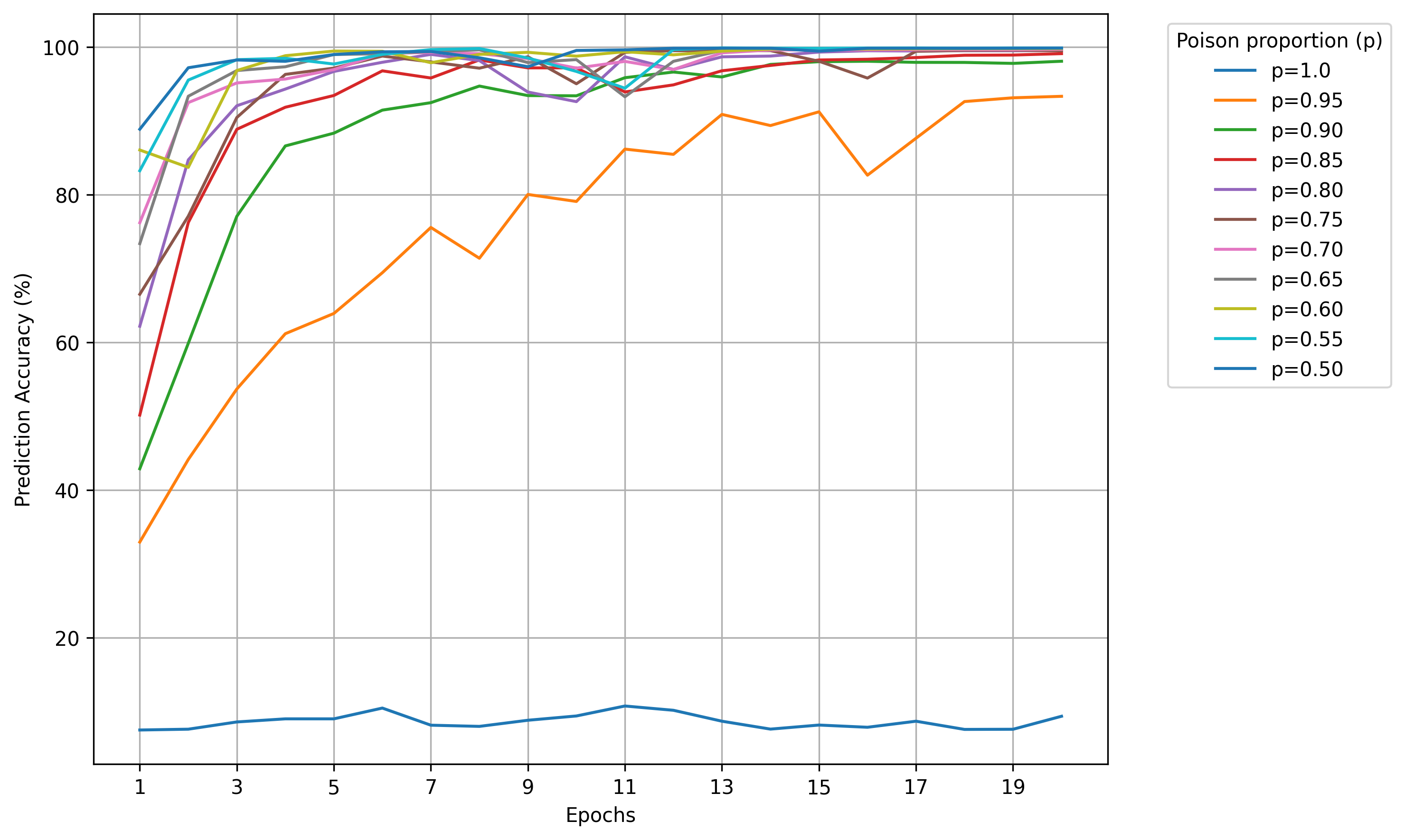}
    \caption{The error-minimizing attacks with different poison proportions.}
    \vspace*{-0.5cm}
    \label{fig:proportions}    
\end{figure}

Next, we demonstrate a limitation of the proposed attack method. Our experiments showed that \authtwo{the} error-minimizing attacks are not effective when only part of the training data is poisoned. To illustrate this, we first poisoned the entire training dataset. Then, we poisoned a proportion of the training dataset and observed the attack's performance. Fig.~\ref{fig:proportions} shows the prediction accuracy curves when poison proportions ($p$) are changed from 1 to 0.5. When the full dataset is poisoned, i.e., $p=1.0$, the prediction accuracy is around 10\%. When only 95\% of the dataset is poisoned, the prediction accuracy is over 90\%. When the poison proportion is further reduced, the model gives even higher prediction accuracies. These results demonstrate that \authtwo{the} error-minimizing attacks are not effective \authtwo{in} fooling traffic sign recognition systems when a portion of the training dataset is clean.

\begin{table}[ht!]
\vspace*{-0.1cm}
    \caption{The prediction accuracy of the traffic sign recognition systems.}
    \label{tab:results}
    \centering
    \begin{tabular}{|p{0.8cm}|p{0.6cm}|p{0.6cm}|p{0.7cm}|p{1.7cm}|p{1.7cm}|}
        \hline
        Dataset &No  &$\epsilon$ & Attack  & Our Mitigation   & Adversarial \\
        &  Attack (\%) & & (\%) &Scheme (\%) & Training (\%)\\
        \hline
        & \multirow{3}{*}{99.90} & 16/255 & 10.6 & 96.05 \textcolor{gray}{(+85.45)} & 91.52 \textcolor{gray}{(+80.92)}  \\ 
        \cline{3-6}
        GTSRB  & & 8/255 & 22.38  & 99.59 \textcolor{gray}{(+77.21)} & 98.87 \textcolor{gray}{(+76.49)}  \\ 
        \cline{3-6}
        \cite{Houben-IJCNN-2013}&  & 4/255 & 52.10   & 99.86 \textcolor{gray}{(+47.76)} & 98.93 \textcolor{gray}{(+46.83)}\\ 
        \cline{3-6}
        \hline 
        &  \multirow{3}{*}{98.56} &16/255 & 25.18 & 98.08 \textcolor{gray}{(+72.90)}& 96.04 \textcolor{gray}{(+70.86)} \\
        \cline{3-6}
        CTSRD  &  & 8/255 & 45.44  & 98.32 \textcolor{gray}{(+52.88)} & 95.20 \textcolor{gray}{(+49.76)}\\
        \cline{3-6}
          ~\cite{tsrd} & & 4/255 & 70.26 & 98.32 \textcolor{gray}{(+28.06)}& 94.96 \textcolor{gray}{(+24.70)} \\ 
        \hline
    \end{tabular}
    \vspace*{-0.3cm}
\end{table}

\subsection{The Detection Model}
\begin{figure*}[htbp!]
   \subfigure[$\epsilon=16/255$]{
        \includegraphics[width=0.3\textwidth]{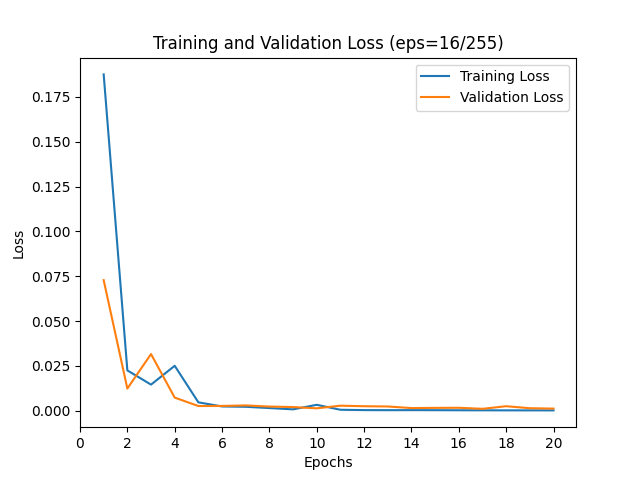}
    }
    \subfigure[$\epsilon=8/255$]{
        \includegraphics[width=0.3\textwidth]{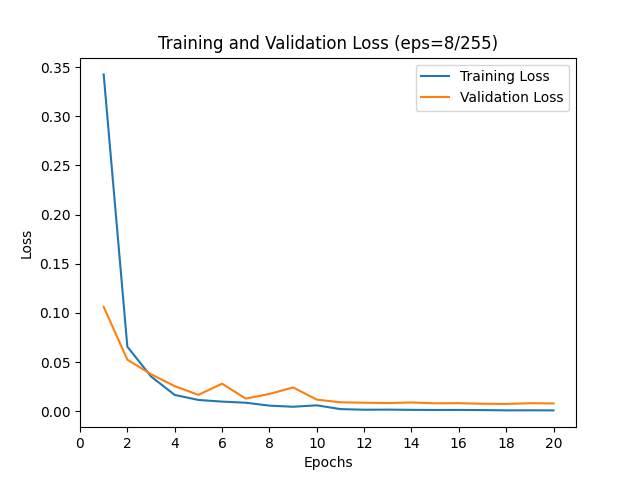}
    }
    \subfigure[$\epsilon=4/255$]{
        \includegraphics[width=0.3\textwidth]{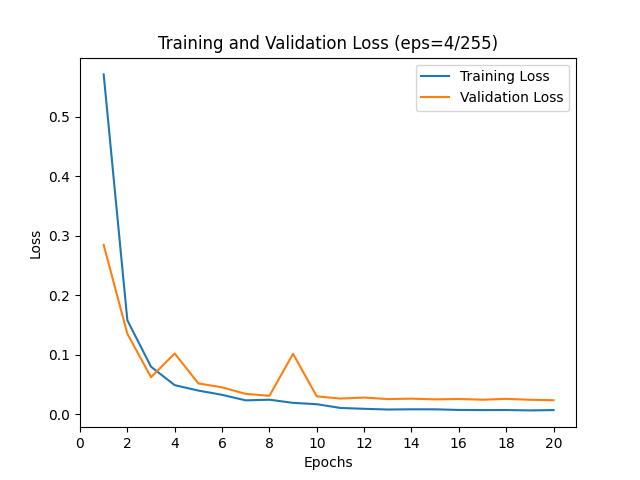}
    }
    \caption{Training and validation loss curves of the detection model.}
    \label{fig:acc_loss_curves}
    \vspace*{-0.4cm}
\end{figure*}

The detection model is evaluated on a version of GTSRB dataset available in Pytorch datasets.  The dataset is divided into training and prediction sets, containing 39,209 images for training and 12,630 images for prediction. We converted 50\% of both training and prediction datasets into poisoned data. The detection model's performance is presented in Table~\ref{tab:detection}. The model achieved a success (prediction accuracy) rate of over 99\% regardless of the strength of the attack ($\epsilon$). These results imply that the detection model can accurately distinguish poisoned data from the clean data. Moreover, the success rate of the detection model \authtwo{is} slightly reduced when the $\epsilon$ is low, \authtwo{probably} due to the lower intensity of the perturbations, making them harder to detect. Fig.~\ref{fig:acc_loss_curves} shows the loss curves for the detection models trained on poisoned data with different $\epsilon$. The detection model trained on data poisoned with a small $\epsilon$ exhibited a higher initial binary cross-entropy loss compared to training with a large $\epsilon$.  However, by the end of training, all detection models reached similar loss values.

\begin{table}[ht]
\vspace{-0.1cm}
\caption{Accuracy of the detection model.}
    \centering
    \begin{tabular}{|c|c|c|}
    \hline
    $\epsilon$ & Training Accuracy & Success Rate \\
    \hline
    16/255 & 100\% & 99.97\% \\
    \hline
    8/255 &  99.99\% & 99.66\% \\
    \hline
    4/255 & 99.80\% & 99.11\% \\
    \hline
    \end{tabular}
    \label{tab:detection}
    \vspace*{-0.5cm}
\end{table}

\subsection{The Mitigation Scheme}

\begin{figure*}[ht!]
    \centering
    \subfigure[GTSRB ($\epsilon=16/255$)]{
        \includegraphics[width=0.3\textwidth]{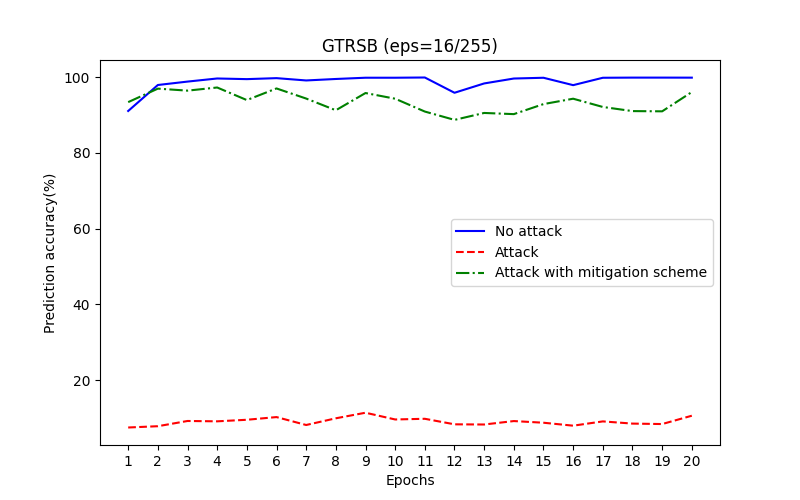}
    }
    \subfigure[GTSRB ($\epsilon=8/255$)]{
        \includegraphics[width=0.3\textwidth]{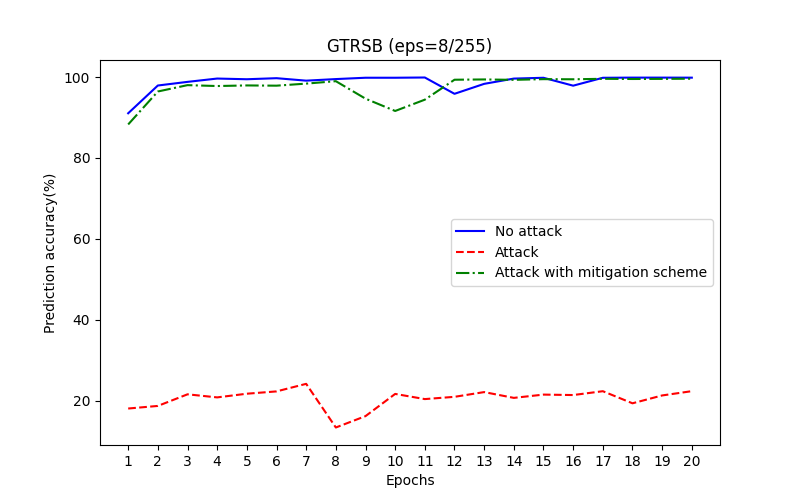}
    }
    \subfigure[GTSRB ($\epsilon=4/255$)]{
        \includegraphics[width=0.3\textwidth]{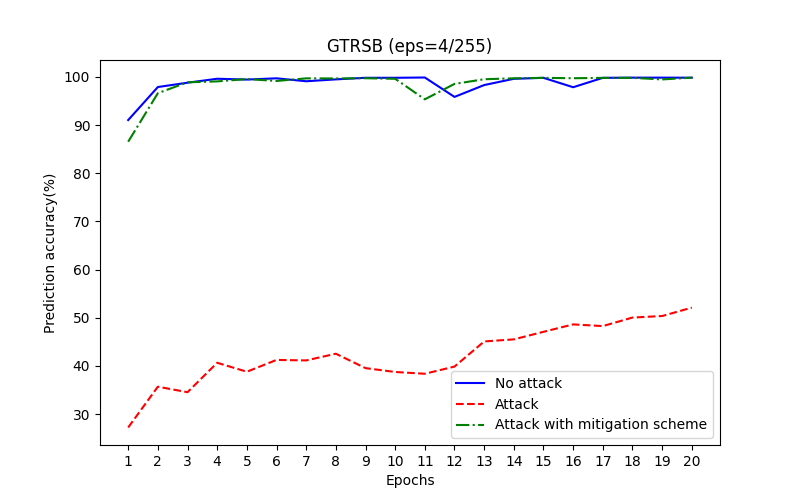}
    }
    \subfigure[CTSRD ($\epsilon=16/255$)]{
        \includegraphics[width=0.3\textwidth]{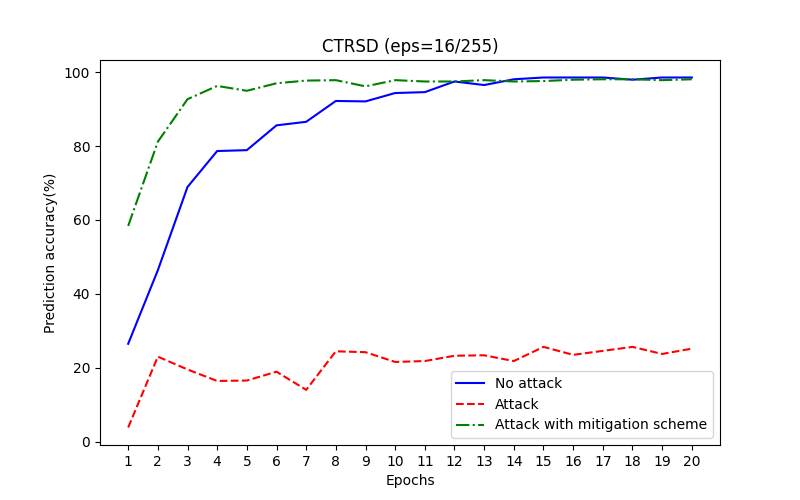}
    }
    \subfigure[CTSRD ($\epsilon=8/255$)]{
        \includegraphics[width=0.3\textwidth]{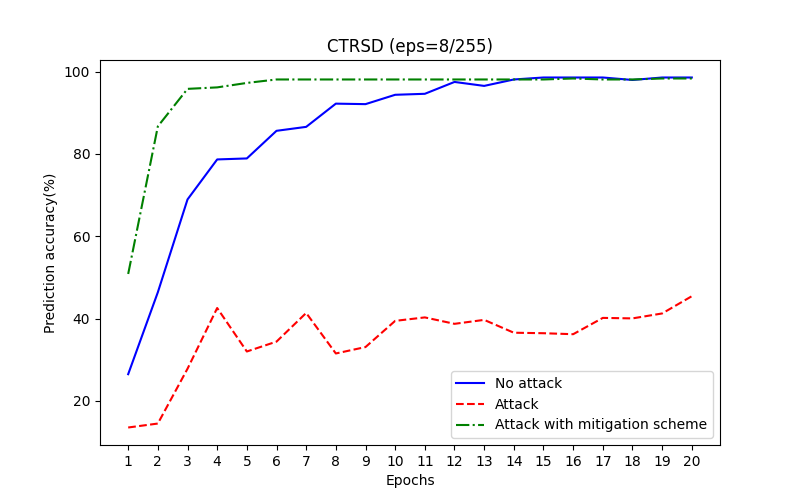}
    }
    \subfigure[CTSRD ($\epsilon=4/255$)]{
        \includegraphics[width=0.3\textwidth]{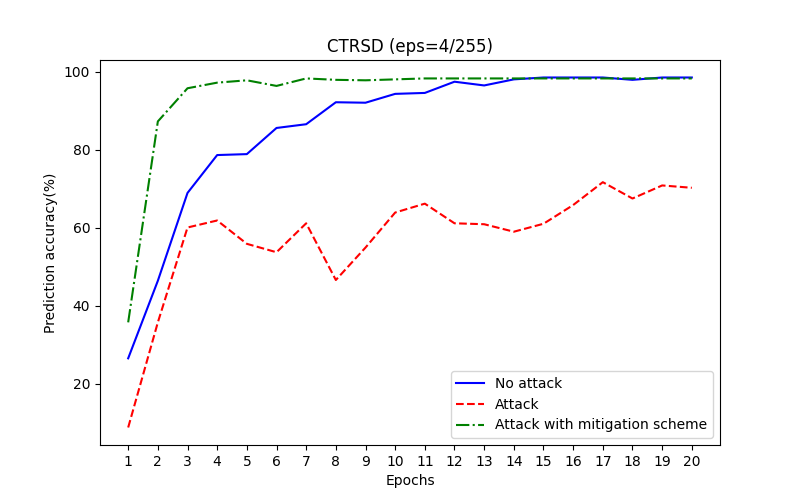}
    }
    \caption{Prediction accuracy during training with and without error-minimization attacks, as well as after the application of our mitigation scheme.}
    \label{fig:prediction_curves}
    \vspace*{-0.4cm}
\end{figure*}

The fifth column of Table~\ref{tab:results} shows the prediction accuracy after applying our mitigation scheme against \authtwo{the} error-minimizing attacks. To mitigate the weaker attacks ($\epsilon$ = 8/255, 4/255) on the GTSRB, we only needed to use grayscale transformation, which increased the prediction accuracy up to 99\%. To mitigate the attack with higher strength ($\epsilon$ = 16/255) on the  GTSRB dataset, we used grayscale, Color Jitter and Random Invert as the nonlinear transformations, improving the prediction accuracy up to 96.05\%. We applied the same transformations to the CTSRD dataset attacked with \authtwo{the} error-minimizing attack at $\epsilon$ = 16/255. It improved the accuracy from 25.18\% to 98.08\%, achieving almost the same prediction accuracy as the model trained on clean data. For mitigating the weaker attacks ($\epsilon$ = 8/255, 4/255) on CTSRD, we used Color Jitter and Grayscale. Fig.~\ref{fig:prediction_curves} shows the evaluation of prediction accuracy during training with and without the error-minimizing attack and after the applying our mitigation scheme. The red lines denote the prediction accuracy when the model is trained with standard training on poisoned data, which is very low. The blue lines show the high prediction accuracy when the same training approach is used on clean data. The green lines indicate the prediction accuracy when the poisoned data is trained using our mitigation scheme, which is almost the same as training with clean data. These results show that the mitigation scheme can overcome the effects of \authtwo{the} error-minimizing attacks and provide \authtwo{a} prediction accuracy nearly equivalent to that of clean data.

We compared our mitigation scheme with adversarial training~\cite{madry2017AT_PGD}, a widely used approach for mitigating evasion and data poisoning attacks. We implemented adversarial training using PGD attacks~\cite{DBLP:conf/iclr/FuHLST22}. Following the default settings in~\cite{DBLP:conf/iclr/FuHLST22}, we used a perturbation radius of 8/255, a step size of 0.8/255, and the number of PGD steps is set to 10. We trained the same ResNet18 model for 20 epochs to be comparable with our other experiments. The prediction accuracies after applying adversarial training are shown in the sixth column of Table~\ref{tab:results}.  Our mitigation scheme outperforms adversarial training regardless of the attack strength. When the attack strength is low ($\epsilon$ = 4/255, 8/255), adversarial training performs as well as our mitigation scheme. However, when the attack strength is high ($\epsilon$ = 16/255), our mitigation scheme performs significantly better than adversarial training. 

\section{Conclusions and Future Work} 

Nowadays, traffic sign recognition systems in autonomous vehicles are predominantly based on DNNs. These DNNs can be compromised through various attacks, including data poisoning. In this paper, we exploited \authtwo{the} error-minimizing attacks to poison DNNs used for traffic sign recognition during training. However, our experiments demonstrated that the attack is effective only when the entire training dataset is poisoned.  
Furthermore, we showed that \authtwo{the} error-minimizing attacks can be mitigated by employing a data-augmentation-based training method. The proposed mitigation scheme was \authtwo{more} effective than the computationally expensive adversarial training approach for mitigating \authtwo{the} error-minimizing attacks. Furthermore, our findings highlighted the necessity of utilizing advanced training techniques for traffic sign recognition systems to enhance their resilience against data poisoning attacks. In the future, we aim to evaluate the robustness of our mitigation scheme on diverse datasets and to consider advanced model architectures in the detection model. We also plan to improve \authtwo{the} error-minimizing attacks by addressing the limitations identified in this study.

\section{Acknowledgment}
We acknowledge NSF for partially sponsoring the work under grants \#1620868 with its REU, \#2228562, and \#2236283. We also thank Cyber Florida for a seed grant.

\bibliographystyle{IEEEtran}
\bibliography{Bibtex}

\end{document}